\documentclass[10pt,twocolumn,letterpaper]{article}

\usepackage[pagenumbers]{cvpr} % To force page numbers, e.g. for an arXiv version
\usepackage[accsupp]{axessibility}
\usepackage{graphicx}
\usepackage{amsmath}
\usepackage{amssymb}
\usepackage{booktabs}
\usepackage{algorithm}
\usepackage{algorithmic}
\usepackage{xspace}
\usepackage{multirow}
\usepackage{booktabs,tabularx}
\usepackage{enumitem}
\usepackage{makecell}
\usepackage{indentfirst}
\usepackage{siunitx}
\usepackage{color}
\usepackage{newfloat}
\usepackage{listings}
\newcommand{\tabincell}[2]{\begin{tabular}{@{}#1@{}}#2\end{tabular}}
\floatstyle{ruled}
\newfloat{listing}{tb}{lst}{}
\floatname{listing}{Listing}
\usepackage[pagebackref,breaklinks,colorlinks]{hyperref}

\usepackage[capitalize]{cleveref}
\crefname{section}{Sec.}{Secs.}
\Crefname{section}{Section}{Sections}
\Crefname{table}{Table}{Tables}
\crefname{table}{Tab.}{Tabs.}

\def\cvprPaperID{8352} % *** Enter the CVPR Paper ID here
\def\confName{CVPR}
\def\confYear{2022}

\begin{document}

\title{Knowledge Mining with Scene Text for Fine-Grained Recognition}
\author{
Hao Wang\textsuperscript{\rm 1}\thanks{Authors contribute equally.},
Junchao Liao\textsuperscript{\rm 1}\footnotemark[\value{footnote}],
Tianheng Cheng\textsuperscript{\rm 1},
Zewen Gao\textsuperscript{\rm 1},
Hao Liu\textsuperscript{\rm 2},
Bo Ren\textsuperscript{\rm 2},
Xiang Bai\textsuperscript{\rm 1},
Wenyu Liu\textsuperscript{\rm 1}\thanks{Corresponding author.}\\
\textsuperscript{\rm 1}Huazhong University of Science and Technology,
\textsuperscript{\rm 2}Tencent YouTu Lab\\
{\tt\small \{wanghao4659,liaojc,thch,gaozw,xbai,liuwy\}@hust.edu.cn, \{ivanhliu,timren\}@tencent.com}\\
}
\maketitle

%%%%%%%%% ABSTRACT
\begin{abstract}
Recently, the semantics of scene text has been proven to be essential in fine-grained image classification. However, the existing methods mainly exploit the literal meaning of scene text for fine-grained recognition, which might be irrelevant when it is not significantly related to objects/scenes. We propose an end-to-end trainable network that mines implicit contextual knowledge behind scene text image and enhance the semantics and correlation to fine-tune the image representation. Unlike the existing methods, our model integrates three modalities: visual feature extraction, text semantics extraction, and correlating background knowledge to fine-grained image classification.  Specifically, we employ KnowBert to retrieve relevant knowledge for semantic representation and combine it with image features for fine-grained classification.
Experiments on two benchmark datasets, Con-Text, and Drink Bottle, show that our method outperforms the state-of-the-art by 3.72\% mAP and 5.39\% mAP, respectively. To further validate the effectiveness of the proposed method, we create a new dataset on crowd activity recognition for the evaluation. The source code and new dataset of this work are available at this repository\footnote{\href{https://github.com/lanfeng4659/KnowledgeMiningWithSceneText}{https://github.com/lanfeng4659/KnowledgeMiningWithSceneText}}.  
\end{abstract}

\section{Introduction}
The text conveys the information, knowledge, and emotion of human beings as a significant carrier. Texts in natural scene images contain sophisticated semantic information that can be used in many vision tasks such as image classification, visual search, and image-based question answering.

Several approaches~\cite{Wang_2021_CVPR,wacv/MaflaDBGK21,cvpr/Movshovitz-Attias15,tmm/KaraogluTGS17,access/BaiYLXL18,wacv/MaflaDBGK20} were proposed to incorporate semantic cues of scene text for image classification or retrieval and achieved significant performance improvements. These methods follow a general pipeline that first spots the text by a scene text reading system, then converts the spotted word into text features to combine it with image features for the subsequent tasks.

\begin{figure}[t]
	\centering
	\includegraphics[width=0.95\linewidth]{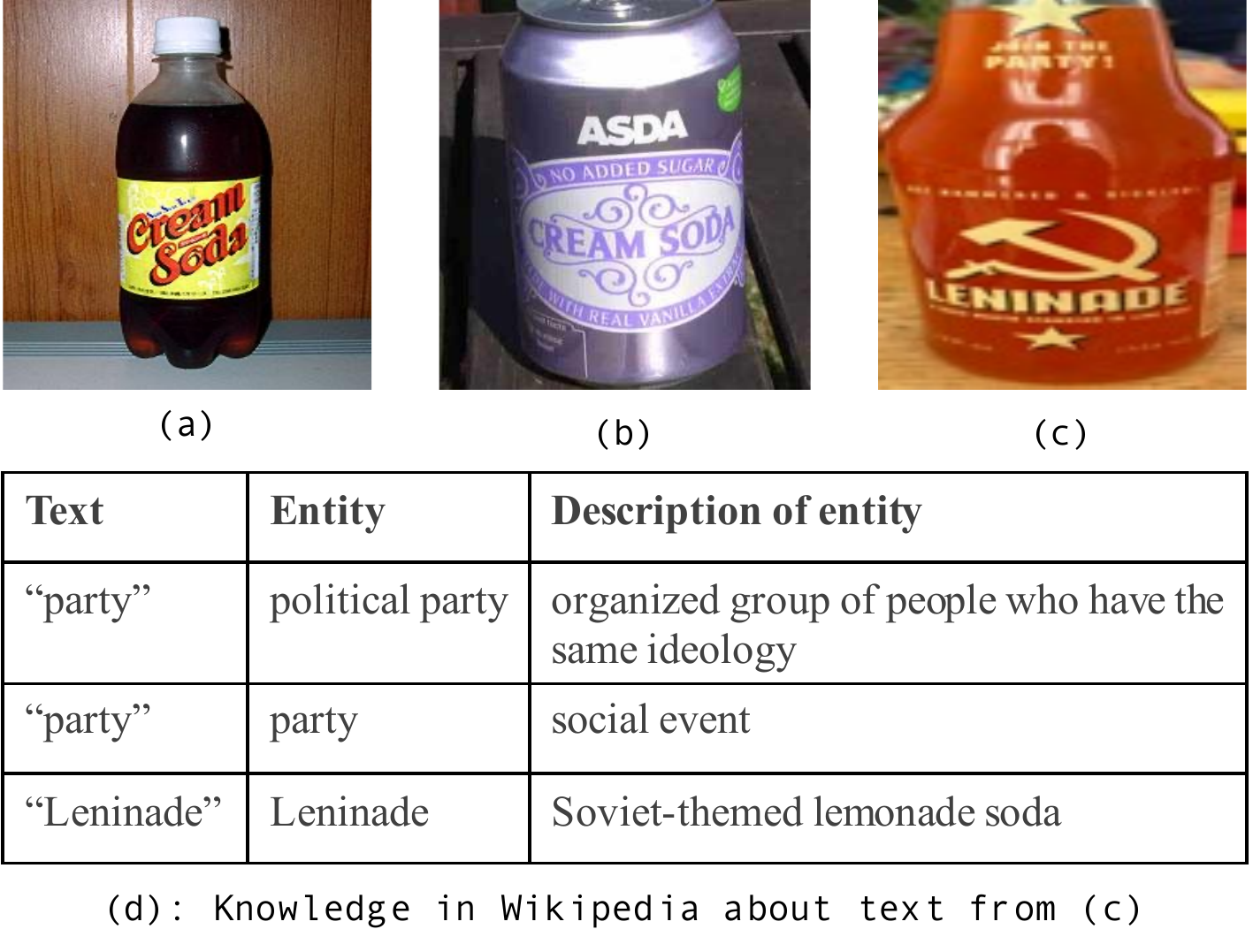}
	\vspace{-1.5ex}
	\caption{The three images belong to the category of ``Soda”. 
	(d) shows the knowledge behind scene text embodied in the image (c) from knowledge base Wikipedia. Each text instance contains one or more entities stored in the knowledge base. The associated descriptions further explain the precise meaning of entity. Only the entities of two text instances are listed for simplicity.}
	\label{motivation}
	  \vspace{-1.5ex}
\end{figure}

This paper explores how to dig deeper into background knowledge and extract context information of scene text for the fine-grained image classification task. Unlike document text, in our observation, the natural scene text is often sparse, appearing as a few keywords rather than complete sentences. Moreover, these few keywords may be vague and give no clue to the classification model when their semantic cues are not directly related to the precise meaning that the image conveys.

As shown in Fig.~\ref{motivation} (a) and (b), the literal meaning of the keyword ``Soda" explicitly expresses that the bottles in the two images belong to the category~\textit{Soda} despite their intra-class visual variance.
However, we hardly understand the object in Fig.~\ref{motivation} (c) by solely fetching the semantic cues of scene text. To understand the image certainly, getting more relevant contextual knowledge about the image is crucial. Therefore, we explore how to dig extra background knowledge and mine the contextual information to enhance the correlation between scene text and a picture. For example, the table in Fig.~\ref{motivation} (d) exhibits related information or knowledge of scene text embodied in (c). The description of the entity~\textit{Leninade} informs that it is a Soda beverage bottle. Thus, the knowledge extracted in this manner complements the literal meaning of the raw text and reduces the semantics loss caused by using the literal meaning of scene text only.

Specifically, after extracting the text from the image by a scene text reading system~\cite{aaai/WangLZYBXHW020,eccv/LiaoPHHB20}, we retrieve relevant knowledge from databases such as (~\, e.g., WordNet~\cite{cacm/Miller95} and Wikipedia) that store rich human-curated knowledge with all possible correlation to the target. 
As shown in Fig.~\ref{motivation} (d), the possible entities (~\, e.g., party and political party) can be extracted for the text instance ``party" from the knowledge databases. However, all the retrieved contextual knowledge may not necessarily provide helpful semantic cues to understand the visual contents. In order to filter relevant contextual information from irrelevant, we design an attention module that focuses on very pertinent knowledge for the semantics of objects or scenes.

We evaluate the performance of our method on two public benchmark datasets, Bottles~\cite{access/BaiYLXL18} and Con-Text~\cite{mm/KaraogluGG13}. The results demonstrate the usage of contextual knowledge behind scene text can significantly promote fine-grained image classification models performances. To further prove the effectiveness of our method, we developed a new dataset consisting of 21 categories and 8785 natural images. Furthermore, the dataset mainly focuses on crowd activity, while most images contain multiple scene text instances. 
To the best of our knowledge, the existing crowd activity datasets do not contain scene text instances. However, everyday human activities are highly related to scene text presences, for example, procession, exhibitions, press briefing, and sales campaigns. This dataset will be a valuable asset for exploring the role of scene text on crowd activity. 

In this paper, we propose a method that mines contextual knowledge behind scene text to improve the performance of the multi-modality understanding task. To this end, we design a deep-learning-based architecture that combines three modality features, including visual contents, scene text, and knowledge for fine-grained image recognition. Our method achieves significant improvements and can be applied to other tasks, such as visual grounding~\cite{iccv/PlummerWCCHL15} and text-visual question answering~\cite{iccv/BitenTMBRJVK19} beyond the fine-grained image classification task. In addition, we propose a new dataset where each image contains multiple scene text instances, which promotes the study of multi-modal crowd activity analysis.

% The remaining parts of this paper are arranged as follows. In Section~\ref{related_work}, we briefly review the methods on fine-grained image classification and knowledge-aware language models. In Section~\ref{method}, we detail the proposed framework that integrates three modality features for the classification task. In Section~\ref{experiment}, we first introduce our proposed crowd activity dataset and evaluate the performances of our method. Finally, we conclude this work and give future research trends based on this work in Section~\ref{conclusion}.

\begin{figure*}[t]
	\centering
	\includegraphics[width=0.95\linewidth]{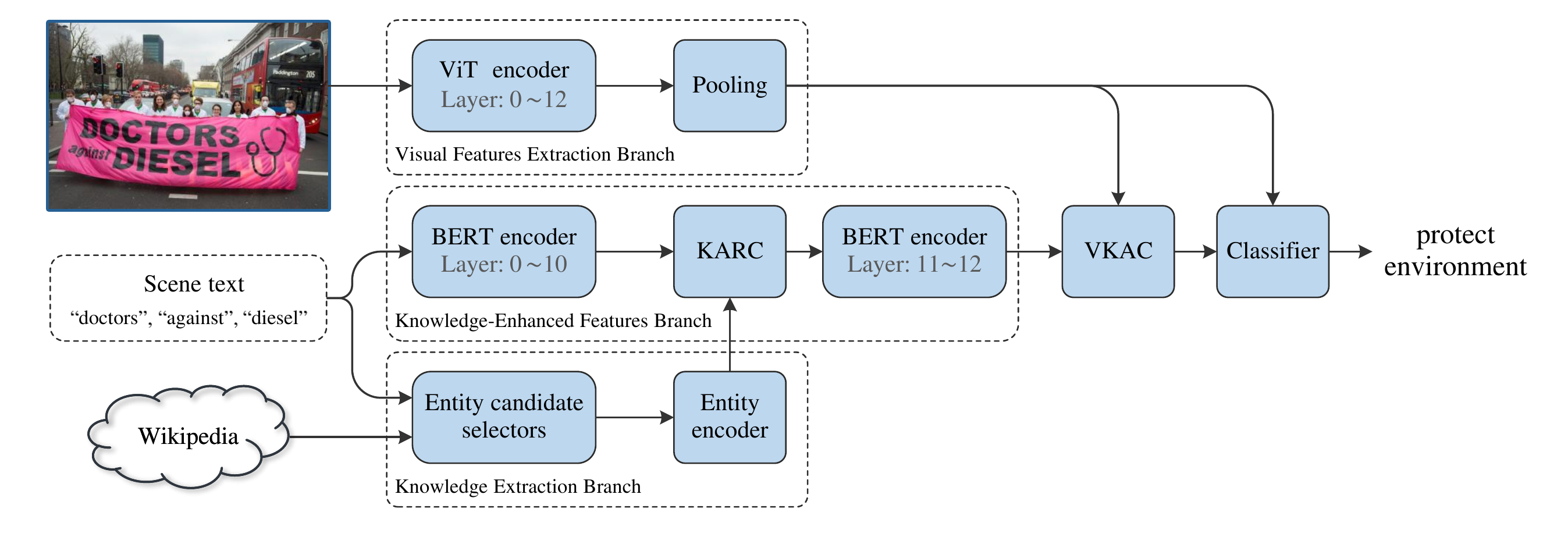}
	% \vspace{-1.5ex}
\caption{The framework of our method. The proposed model combines visual cues and textual cues for classification. The input text instances are spotted by a scene text reading system.
KARC and VKAC mean the knowledge attention and recontextualization component and the visual-knowledge attention component, respectively.}
\label{framework}
%   \vspace{-1.5ex}
\end{figure*}

\section{Related Work}\label{related_work}

\subsection{Fine-Grained Image Classification}
The task of fine-grained image classification needs to distinguish images with subtle visual differences among object classes in some domains, such as animal species~\cite{dicta/GeMSWLRC16,khosla2011novel}, plant species~\cite{corr/MajiRKBV13} and man-made objects~\cite{iccvw/Krause0DF13}. 
Previous methods~\cite{cvpr/FuZM17,Cordonnier2021CVPR} classify objects with only visual cues and aim at finding a discriminative image path.
Recently, some approaches have shown a growing interest in employing textual cues to combine the visual cues for this task.
Movshovitz~\etal~\cite{cvpr/Movshovitz-Attias15} first propose to leverage scene text for the fine-grained image classification task by using the visual cues of scene text.  
However, extracting robust visual cues of scene text is challenging due to blur and occlusion of text instances.
Karaoglu~\etal~\cite{tmm/KaraogluTGS17} employ the textual cues of scene text as a discriminative signal and combine the visual features that are obtained by the GoogLeNet~\cite{cvpr/SzegedyLJSRAEVR15} to distinguish business place.
To fully exploit the complementarity of visual information and textual cues, several methods~\cite{access/BaiYLXL18,wacv/MaflaDBGK20} propose to fuse features of the two modalities with an attentional module.
Bai~\etal~\cite{access/BaiYLXL18} propose an attention mechanism to select textual features from word embeddings of recognized words. 
To overcome optical character recognition errors, Mafla~\etal~\cite{wacv/MaflaDBGK20} leverage the usage of the PHOC~\cite{pami/AlmazanGFV14} representation to construct a bag of textual words along with the fisher vector~\cite{cvpr/PerronninD07} that models the morphology of text.
Despite the promising progress, the existing methods exploit the literal meaning of scene text and overlook the meaningful human-curated knowledge of text.

\subsection{Knowledge-aware Language Models}
The pre-trained language models such as ELMo~\cite{naacl/PetersNIGCLZ18} and BERT~\cite{naacl/DevlinCLT19} are optimized to either predict the next word or some masked words in a given sequence.
Petroni~\etal~\cite{petroni2019language} find that the pre-trained language models, such as BERT, can recall factual and commonsense knowledge. Such knowledge is stored implicitly in the parameters of the language model and useful for downstream tasks such as visual question answering~\cite{emnlp/KhanMLS20}. 
This knowledge is usually obtained either from the latent context representations produced by the pre-trained model or by using the parameters of the pre-trained model to initialize a task-specific model for further fine-tuning.
To further enhance the language model awareness of human-curated knowledge better, some works~\cite{Peters2019KnowledgeEC,corr/abs-2007-00655} explicitly integrate the knowledge in knowledge bases into the pre-trained language model.
In our method, we employ both BERT~\cite{naacl/DevlinCLT19} and KnowBert~\cite{Peters2019KnowledgeEC} as a knowledge-aware language model and apply them to extract knowledge features.
Although previous methods~\cite{textvqaknow} extract knowledge features from sentences on vision-language tasks, they require the annotation of image-text pairs.

\section{Methodology}\label{method}
As shown in Fig.~\ref{framework}, the proposed network accepts as input an image, a knowledge base, and scene text spotted by a scene text reading system such as~\cite{aaai/WangLZYBXHW020,eccv/LiaoPHHB20}. The part of extracting features in our framework consists of three branches, the visual features extraction branch, the knowledge extraction branch for retrieving relevant knowledge, and the knowledge-enhanced features branch that employs the retrieved knowledge to enhance the presentations of scene text. 
Then, the visual-knowledge attention component (VKAC) inputs the visual features and the knowledge-enhanced text features and outputs the attended features.
Moreover, the concatenation of visual features and attended features is fed to the subsequent classifier.

In our method, we employ ViT~\cite{iclr/DosovitskiyB0WZ21} to extract the global visual features of the input image. We mainly detail the knowledge extraction branch, Knowledge-enhanced features branch, and the visual-knowledge attention component in the following subsections.
% The visual features are the outputs of the encoder of Vision Transformer (ViT)~\cite{iclr/DosovitskiyB0WZ21}.

\subsection{Knowledge extraction branch}
The goal of this branch is to extract relevant knowledge from Wikipedia and embed them into features. Such knowledge is stored via entities in a knowledge base, and relevant entities can be queried by scene text instances in our method. However, most text instances can map to multiple entities due to the uncertainty of the meaning of the text. For example, the text ``apple'' can denote the entity of either fruit apple or Apple company. This requires an entity candidate selector that takes as input a sentence and returns a list of $C$ potential entities. 

\begin{figure}[t]
	\centering
	\includegraphics[width=0.98\linewidth]{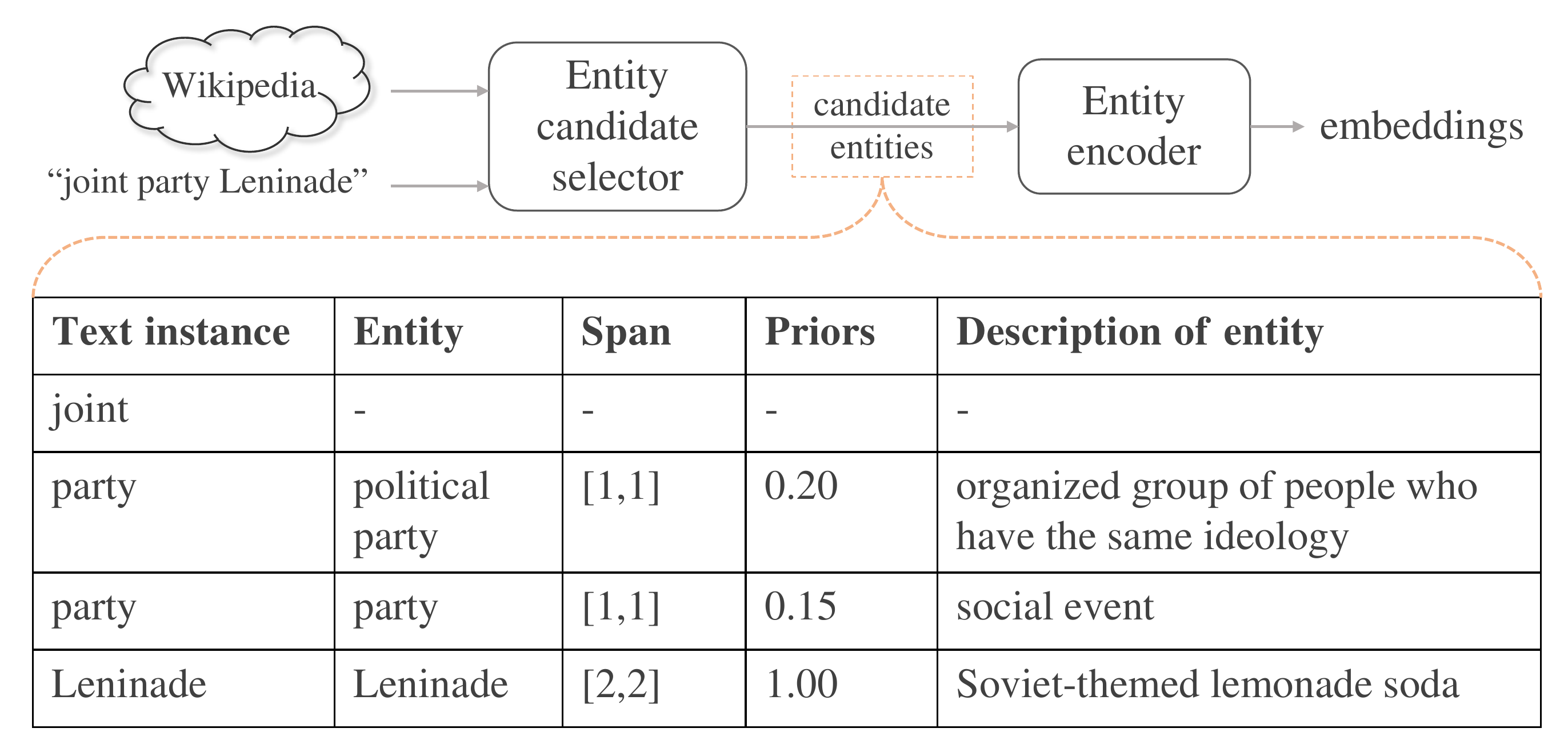}
	\vspace{-1.5ex}
\caption{The process of knowledge extraction branch. The span is the [start index, end index] of the token inside the sentence.}
\label{candidates}
  \vspace{-1.5ex}
\end{figure}

Inspired by~\cite{emnlp/GaneaH17}, we use an entity prior for entity candidate selection. The prior means the probability of a text instance being an entity, which is computed by averaging hyperlink count statistics from Wikipedia, a large Web corpus~\cite{lrec/SpitkovskyC12}, and the YAGO dictionary~\cite{emnlp/HoffartYBFPSTTW11}. As depicted in Fig.~\ref{candidates}, first, we combine all scene text instances to sentence according to the spotting order. Then, the tokens of this sentence are obtained as BERT does. The entity candidate selector generates the top $C$ entity candidates of each text instance based on the prior. 
% Meanwhile, the Wikipedia description of each entity is fetched. 
Finally, the entity embeddings are obtained via the precomputed entity encoder in KnowBert. Specifically, the entity encoder adopts a skip-gram like objective to learn 300-dimensional embeddings of Wikipedia page titles from Wikipedia descriptions. As a result, such entity embeddings encode the factual knowledge mined from Wikipedia descriptions.

\subsection{Knowledge-enhanced features branch}

\begin{figure}[t]
	\centering
	\includegraphics[width=0.98\linewidth]{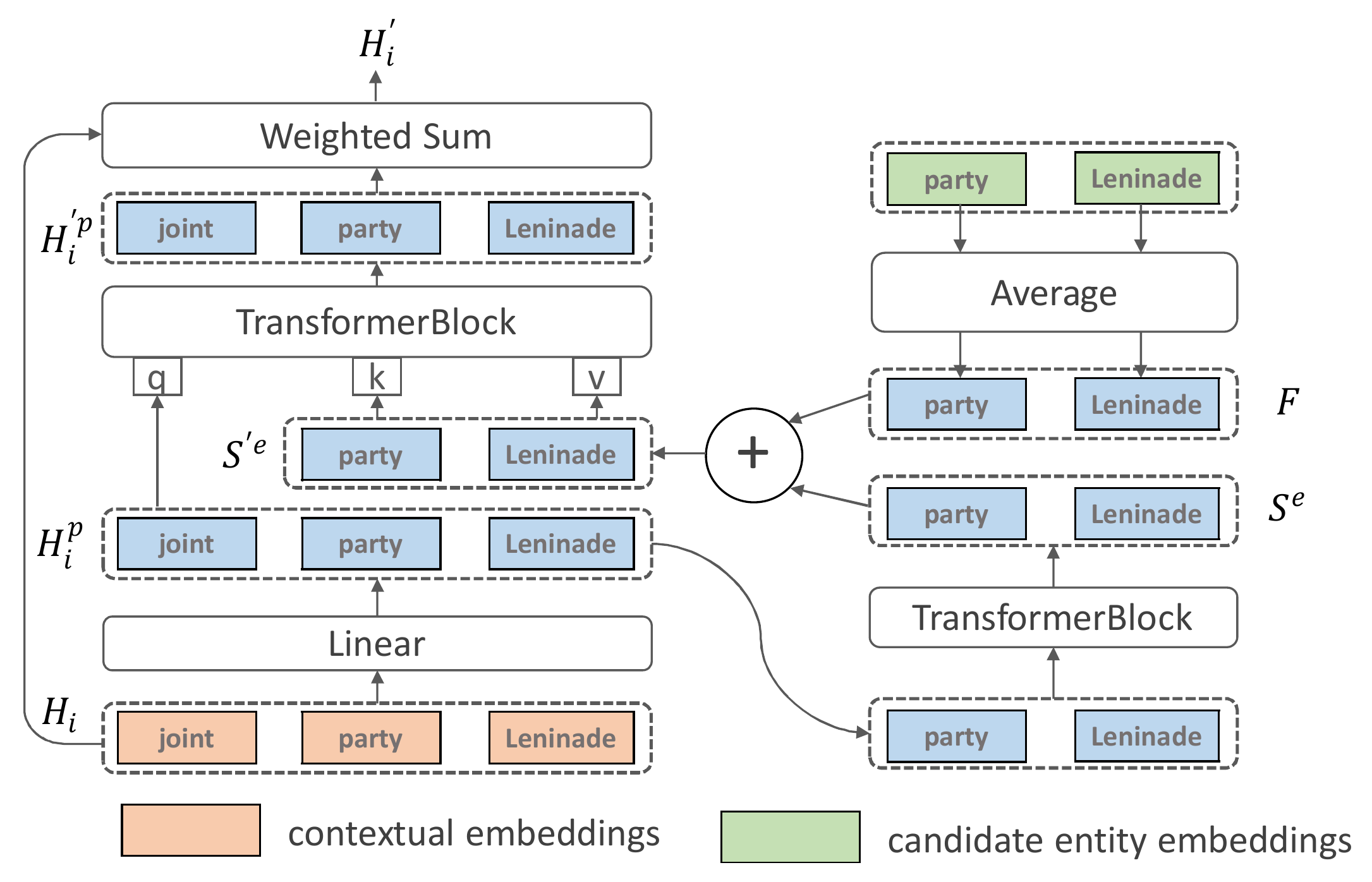}
	\vspace{-1.5ex}
\caption{The architecture of the knowledge attention and recontextualization component.}
\label{karc}
  \vspace{-1.5ex}
\end{figure}

This branch aims at using the retrieved entity embeddings to enhance the representations of text. The architecture is adapted from KnowBert that incorporates knowledge bases into BERT by inserting a knowledge attention and recontextualization component (KARC) at a particular layer. Following KnowBert, we insert Wikipedia into the $10^{th}$ layer of the encoder of BERT.

The brief pipeline of this branch is given in Fig.~\ref{framework}. Formally, a sequence of word piece tokens is fed to the former 10 successive encoder layers of BERT, outputting the contextual representations $H_i$. Then, the KARC takes as inputs $H_i$ and candidate entity embeddings and outputs knowledge enhanced representations $H_i^{'}$. Finally, these enhanced representations are fed to the remainder of the encoder of BERT, generating the final knowledge enhanced features.
The module in each encoder layer of BERT is the TransformerBlock formulated as
\begin{equation}
    H_i = \text{TransformerBlock}(H_{i-1}, H_{i-1}, H_{i-1}).
\end{equation}
This block uses $H_{i-1}$ as the query, key, and value to allow each vector to attend to each other.

\begin{figure*}[t]
    \includegraphics[width=0.98\linewidth]{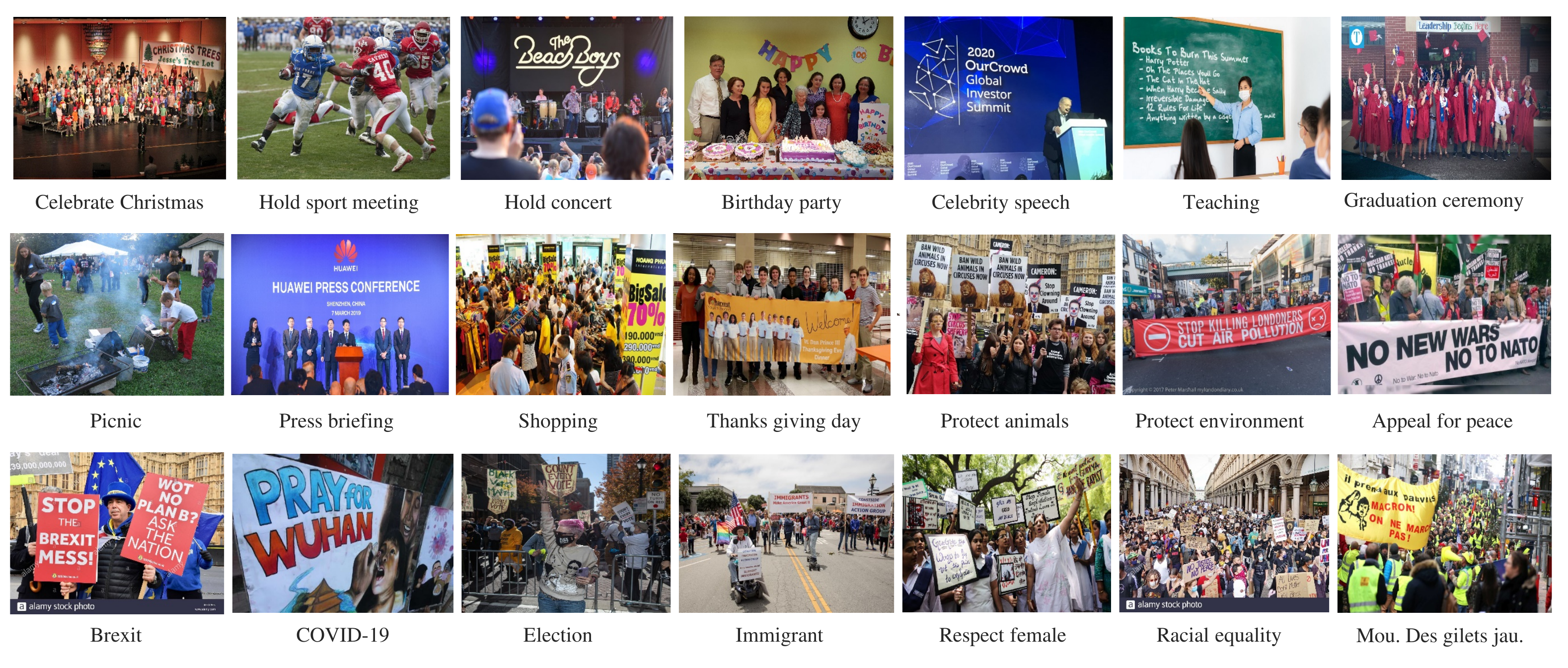}
\centering
\vspace{-1.5ex}
  \caption{Examples of 21 categories from the Crowd Activity dataset.}
  \label{examples}
  \vspace{-1.5ex}
\end{figure*}

\begin{table*}[h]
\setlength\tabcolsep{1pt}
\renewcommand{\arraystretch}{1.2}
\centering  % 表居中
\begin{tabular}{c | c c c c c c c c c c c | c c c c c c c c c c | c}
\hline
\multirow{2}*{\small{Method}}& \multicolumn{11}{c|}{\small{Activities of daily living}} & \multicolumn{10}{c|}{\small{Demonstrations}}& \multirow{2}*{\small{mAP}} \\
\cline{2-22}
 &\small{~c.c.} &\small{~h.s.~} &\small{~h.c.~} &\small{~b.p.~} &\small{~c.s.~} &\small{~teac.} &\small{~g.c.~} &\small{~pic.} &\small{~p.b.~} &\small{~shop.} &\small{~t.g.~} &\small{~p.a.~} &\small{~p.e.~} &\small{~a.p.~} &\small{~brex.} &\small{~cov.} &\small{~elec.} &\small{~imm.} &\small{~r.f.~} &\small{~r.e.~} &\small{~m.d.~} & \\ 
\hline
\small{R152~\cite{cvpr/HeZRS16}} & \scriptsize{59.6} & \scriptsize{92.5} & \scriptsize{70.4} & \scriptsize{71.1} & \scriptsize{48.4} & \scriptsize{88.7} & \scriptsize{89.6} & \scriptsize{80.5} & \scriptsize{83.2} & \scriptsize{86.4} & \scriptsize{68.8} & \scriptsize{68.5} & \scriptsize{62.4} & \scriptsize{74.3} & \scriptsize{84.7} & \scriptsize{50.6} & \scriptsize{59.6} & \scriptsize{56.5} & \scriptsize{68.8} & \scriptsize{48.2} & \scriptsize{91.0} & \scriptsize{71.6} \\
% \hline
\small{ViT~\cite{iclr/DosovitskiyB0WZ21}} & \scriptsize{76.0} & \scriptsize{\textbf{98.7}} & \scriptsize{81.1} & \scriptsize{83.6} & \scriptsize{57.1} & \scriptsize{85.3} & \scriptsize{93.5} & \scriptsize{84.7} & \scriptsize{93.0} & \scriptsize{88.6} & \scriptsize{73.6} & \scriptsize{73.6} & \scriptsize{71.5} & \scriptsize{78.8} & \scriptsize{85.7} & \scriptsize{75.6} & \scriptsize{74.6} & \scriptsize{75.8} & \scriptsize{84.3} & \scriptsize{60.0} & \scriptsize{91.0} & \scriptsize{80.3} \\ 
\hline
\small{fastText~\cite{bojanowski2017enriching}} & \scriptsize{58.3} & \scriptsize{46.9} & \scriptsize{55.3} & \scriptsize{56.0} & \scriptsize{33.4} & \scriptsize{46.1} & \scriptsize{59.6} & \scriptsize{31.7} & \scriptsize{52.0} & \scriptsize{47.9} & \scriptsize{27.1} & \scriptsize{87.2} & \scriptsize{82.7} & \scriptsize{76.7} & \scriptsize{78.9} & \scriptsize{57.6} & \scriptsize{69.6} & \scriptsize{69.8} & \scriptsize{73.0} & \scriptsize{55.0} & \scriptsize{75.4} & \scriptsize{59.1} \\ 
% \hline
\small{KB~\cite{Peters2019KnowledgeEC}} & \scriptsize{62.8} & \scriptsize{56.3} & \scriptsize{55.3} & \scriptsize{59.5} & \scriptsize{51.4} & \scriptsize{54.1} & \scriptsize{70.3} & \scriptsize{46.1} & \scriptsize{54.8} & \scriptsize{45.9} & \scriptsize{45.8} & \scriptsize{89.9} & \scriptsize{79.2} & \scriptsize{78.5} & \scriptsize{78.1} & \scriptsize{72.8} & \scriptsize{73.8} & \scriptsize{67.0} & \scriptsize{77.2} & \scriptsize{64.2} & \scriptsize{74.9} & \scriptsize{64.7} \\ 
\hline
\small{Mafla~\etal~\cite{wacv/MaflaDBGK20}} & \scriptsize{60.0} & \scriptsize{90.9} & \scriptsize{75.6} & \scriptsize{76.4} & \scriptsize{49.4} & \scriptsize{89.0} & \scriptsize{86.4} & \scriptsize{83.5} & \scriptsize{79.0} & \scriptsize{94.2} & \scriptsize{67.1} & \scriptsize{83.1} & \scriptsize{76.2} & \scriptsize{82.4} & \scriptsize{88.7} & \scriptsize{65.5} & \scriptsize{72.4} & \scriptsize{71.4} & \scriptsize{74.7} & \scriptsize{67.7} & \scriptsize{95.5} & \scriptsize{77.6} \\ 
% \hline
\small{Mafla~\etal~\cite{wacv/MaflaDBGK21}} & \scriptsize{72.3} & \scriptsize{87.5} & \scriptsize{78.1} & \scriptsize{80.7} & \scriptsize{50.3} & \scriptsize{91.6} & \scriptsize{86.5} & \scriptsize{81.1} & \scriptsize{73.0} & \scriptsize{89.1} & \scriptsize{62.6} & \scriptsize{87.4} & \scriptsize{79.2} & \scriptsize{86.5} & \scriptsize{85.6} & \scriptsize{75.5} & \scriptsize{79.1} & \scriptsize{73.1} & \scriptsize{80.3} & \scriptsize{67.9} & \scriptsize{97.0} & \scriptsize{79.2} \\ 
\hline
% \small{\textbf{Ours}} & \scriptsize{\textbf{80.6}} & \scriptsize{\textbf{98.4}} & \scriptsize{\textbf{85.9}} & \scriptsize{\textbf{86.3}} & \scriptsize{\textbf{62.5}} & \scriptsize{\textbf{86.7}} & \scriptsize{\textbf{96.6}} & \scriptsize{\textbf{85.8}} & \scriptsize{\textbf{92.7}} & \scriptsize{\textbf{92.8}} & \scriptsize{\textbf{78.0}} & \scriptsize{\textbf{91.4}} & \scriptsize{\textbf{91.7}} & \scriptsize{\textbf{90.8}} & \scriptsize{\textbf{93.2}} & \scriptsize{\textbf{77.6}} & \scriptsize{\textbf{83.1}} & \scriptsize{\textbf{85.4}} & \scriptsize{\textbf{88.7}} & \scriptsize{\textbf{72.8}} & \scriptsize{\textbf{98.0}} & \scriptsize{\textbf{86.6}} \\ 
\small{\textbf{Ours}} & \scriptsize{\textbf{83.0}} & \scriptsize{98.5} & \scriptsize{\textbf{88.8}} & \scriptsize{\textbf{86.1}} & \scriptsize{\textbf{60.5}} & \scriptsize{\textbf{89.4}} & \scriptsize{\textbf{95.7}} & \scriptsize{\textbf{89.1}} & \scriptsize{\textbf{94.0}} & \scriptsize{\textbf{94.5}} & \scriptsize{\textbf{78.2}} & \scriptsize{\textbf{92.4}} & \scriptsize{\textbf{92.4}} & \scriptsize{\textbf{89.6}} & \scriptsize{\textbf{95.4}} & \scriptsize{\textbf{83.0}} & \scriptsize{\textbf{82.1}} & \scriptsize{\textbf{84.7}} & \scriptsize{\textbf{90.1}} & \scriptsize{\textbf{73.7}} & \scriptsize{\textbf{98.1}} & \scriptsize{\textbf{87.5}} \\ 
\cline{2-23}
\small{\textbf{Gain}} & \multicolumn{11}{c|}{\scriptsize{\textbf{3.9}}} & \multicolumn{10}{c|}{\scriptsize{\textbf{11.1}}}& \scriptsize{\textbf{7.2}}\\ 
\hline
\end{tabular}
%\captionsetup{justification=centering}
\caption{Classification performance for baselines and the proposed method on the Crowd Activity dataset. KB denotes KnowBert.}
\label{table:baselines on crowd activity}
\vspace{-1.5ex}
\end{table*}

The KARC is the key component for integrating the retrieved entity embeddings to $H_i$. Different from the one in KnowBert, the width of the span is restricted as 1 in our KARC. Namely, these entities named as more than one text instance are ignored due to the sparsity of scene text.
The details of KARC are given in Fig.~\ref{karc}, the word piece representations ($H_i$) are first projected to $H_i^p$ by a linear layer. The representations of those word pieces that link to at least one entity are contextualized into contextual word representations $S^e$ by a TransformerBlock. Meanwhile, the $C$ candidate entity representations of each token are averaged to form weighted entity embeddings $F$. Specifically, as KnowBert does, we disregard all candidate entities with scores below a fixed threshold, and softmax normalize the remaining scores to weight the corresponding candidate entity representations. Then, $S^e$ are updated by adding entity embeddings $F$ to form word-entity representations $S^{'e}$.
The $S^{'e}$ is employed to recontextualize the $H_i^p$ with a TransformerBlock, where we substitute $H_i^p$ for the query, and $S^{'e}$ for both the key and value:
\begin{equation}
    H_i^{'p} = \text{TransformerBlock}(H_i^p, S^{'e}, S^{'e}),
\end{equation}
Finally, a residual connection is adapted to fuse the $H_i^{'p}$ and $H_i$, forming the knowledge enhanced representations $H_i^{'}$:
\begin{equation}
    H_i^{'} = g(H_i^{'p}) + H_i,
\end{equation}
where, $g$ is a linear function. The fully connected layer is employed in our method.

\subsection{Visual-knowledge attention component}
Generally, not all knowledge of text in an image must have semantic relations to the object or scene. Some retrieved knowledge may have strong correlations with the image, others may be not relevant at all. Therefore, we design an attention component that focuses on very pertinent knowledge for the semantics of objects or scene. The basic idea is that we take the global visual feature $f_v \in \mathbb{R}^{1\times D}$ as query and retrieve those knowledge features that are highly similar to $f_v$ from all knowledge features $H \in \mathbb{R}^{N\times D}$. The parameter $D$ is the feature dimension.

Formally, given $f_v$ and $H$, we first calculate their similarities, which is defined by:
\begin{equation}
    W = softmax(\frac{\theta(f_v)\cdot(\phi(H))^{T}}{\sqrt{D}}),
\end{equation}
where both $\theta$ and $\phi$ are a single linear function that projects the features into a feature space, $W \in \mathbb{R}^{1\times N}$ is the outputted similarity matrix. Then, $W$ is used for weighting knowledge features. Finally, the weighted features are fed to a residual connection block. The implemented process is defined as follow:
\begin{equation}
    H_{att} = W\cdot\psi(H),
\end{equation}
\begin{equation}
    H_{out} = \kappa(H_{att}) + H_{att},
\end{equation}
where $H_{out} \in \mathbb{R}^{1\times D}$ is the attended knowledge features, $\kappa$ is a linear function.

\subsection{Classifier and loss function}
The classifier consisting of a fully connected layer and a softmax layer performs the classification task, inputting the concatenation of the global visual features and the knowledge-enhanced features.
The objective function is formulated as 
\begin{equation}
L = -\frac{1}{M}\sum_{m=1}^{M}\textbf{1}(m=y)\log{p_m},
\end{equation}
where $M$ is the number of categories, $p_m$ is the probability of predicting the sample as the $m^{th}$ category, $y$ is the associated label.

\begin{table*}[t]
	\centering
	\begin{tabular}{m{.3\columnwidth}|
		m{.3\columnwidth}m{.3\columnwidth}m{.2\columnwidth}|
		m{.16\columnwidth}<{\centering}m{.16\columnwidth}<{\centering}m{.16\columnwidth}<{\centering}}
	\Xhline{1.0pt}
	Method& Vision& Text Spotter& Embedding& Con-Text& Bottles& Activity\\
	\hline
	Karao.~\etal~\cite{mm/KaraogluGG13}& BOW& Custom& BoB& 39.00& -& -\\
	% \hline
	Karao.~\etal~\cite{tmm/KaraogluTGS17}& BOW+GoogLeNet& Jaderberg& Probs& 77.30& -& -\\
	Bai~\etal~\cite{access/BaiYLXL18}& GoogLeNet& Textboxes& GloVe& 78.90& -& -\\
% 	Bai$^\dagger$~\etal~\cite{access/BaiYLXL18}& GoogLeNet& Textboxes& GloVe& 79.60& 72.80& -\\
	Bai$^\dagger$~\etal~\cite{access/BaiYLXL18}& GoogLeNet& Google OCR& GloVe& 80.50& 74.50& -\\
% 	Mafla~\small{\cite{wacv/MaflaDBGK20}}& ResNet-152& TextSpotter& fastText& 74.33& 75.25& -\\
	Mafla~\etal~\cite{wacv/MaflaDBGK20}& ResNet-152& E2E-MLT& GloVe& 77.58& 74.91& 72.58\\
	Mafla~\etal~\cite{wacv/MaflaDBGK20}& ResNet-152& E2E-MLT& fastText& 77.77& 75.40& 73.01\\
	Mafla~\etal~\cite{wacv/MaflaDBGK20}& ResNet-152& SSTR-PHOC& PHOC& 77.45& 75.93& 73.84\\
	Mafla~\etal~\cite{wacv/MaflaDBGK20}& ResNet-152& SSTR-PHOC& FV& 80.21& 77.38& 77.57\\
	Mafla~\etal~\cite{wacv/MaflaDBGK21}& ResNet-152& E2E-MLT& fastText& 82.36& 78.14&75.31\\
	Mafla~\etal~\cite{wacv/MaflaDBGK21}& ResNet-152& SSTR-PHOC& PHOC& 82.77& 78.27&75.45\\
	Mafla~\etal~\cite{wacv/MaflaDBGK21}& ResNet-152& SSTR-PHOC& FV& 83.15& 77.86&77.54\\
	Mafla~\etal~\cite{wacv/MaflaDBGK21}& ResNet-152& Google OCR& fastText& 85.81& 79.87& 79.25\\
	\hline
	Ours& ResNet-152& E2E-MLT& KnowBert& 84.93& 79.32& 81.91\\
    Ours& ViT& E2E-MLT& KnowBert& 87.28& 84.01& 85.68\\
	\textbf{Ours}  & ViT& Google OCR& KnowBert& \textbf{89.53}& \textbf{85.26} & \textbf{87.45}\\
	\Xhline{1.0pt}
	\end{tabular}
% 	\vspace{+0.8ex}
	\caption{Classification performance of state-of-the-art methods on the Con-Text, Drink-Bottle, and Activity datasets. BOW denotes bag of visual words. BoB denotes Bag of Bigrams. FV denotes Fisher Vector.}
	\label{sota_result}
	\vspace{-1.5ex}
\end{table*}

\section{Experiments}\label{experiment}
First, we introduce the datasets used in our experiments and the new dataset created by us. Then, the implementation details are given. Third, we evaluate our method on our proposed Crowd Activity dataset and make comparisons with the state-of-the-art approaches. Last, we conduct the ablation studies. We compare with previous methods under the metric of mAP as most existing methods do. 
% In addition, we provide the classification accuracy in the ablation study subsection.

\subsection{Datasets}
% As for \textbf{evaluation metrics}, we adopt mAP from \cite{wacv/MaflaDBGK20} as metric on all three datasets. Besides, we also calculate the Accuracy, as shown in Tab. \ref{ablation_acc}.

\textbf{Con-Text} dataset is introduced by Karaoglu~\cite{mm/KaraogluGG13} and is a subset of ImageNet dataset~\cite{cvpr/DengDSLL009}. This dataset is constructed by selecting the sub-categories of ``building" and ``place of business", consisting of 24,255 images classified into 28 categories that are visually similar.

\textbf{Drink Bottle} dataset is presented by Bai~\cite{access/BaiYLXL18} and consists of various types of drink bottle images contained in soft drink and alcoholic drink sets in ImageNet dataset~\cite{cvpr/DengDSLL009}. The dataset has 18,488 images divided into 20 categories.

All categories within the existing two datasets are about products or places of business. The textual cues of those categories are obvious, and most images can be understood by the apparent meaning of scene texts rather than the knowledge behind them. Therefore, we create a new dataset that concentrates on the activities of the crowd for a fine-grained image classification task, named as~\textbf{Crowd Activity} dataset, as automatically understanding crowd activity is meaningful for social security.
This dataset is newly collected, where the images are mainly searched on the Internet and collected from streets by mobile phones. All images in this dataset contain at least one text instances. The categories come from activities of daily living and demonstrations stimulated by hot events in recent years. Specifically, this dataset consists of 21 categories and 8785 images in total. As shown in Fig.~\ref{examples}, the 21 categories broadly fall into two types: activities of daily living(~\ie, \textit{celebrating Christmas},~\textit{holding sport meeting,}~\textit{holding concert},~\textit{celebrating birthday party},~\textit{celebrity speech},~\textit{teaching},~\textit{graduation ceremony},~\textit{picnic},~\textit{press briefing},~\textit{shopping},~\textit{celebrating Thanks giving day}) and demonstrations (~\ie, \textit{protecting animals},~\textit{protecting environment},~\textit{appealing for peace},~\textit{Brexit},~\textit{COVID-19},~\textit{election},~\textit{immigrant},~\textit{respecting female},~\textit{racial equality},~\textit{mouvement des gilets jaunes}).

% , persons of different nations, ages, and organizations usually express the same intent with different words.
\subsection{Implementation Details}
Before training, we first extract scene text by Google OCR or E2E-MLT. Then, the model of our method is trained in an end-to-end manner. For the data augmentation on images, we first randomly crop an image patch on the original image with the scale from 0.05 to 1.0 while keeping the ratio in a range of [0.75, 1.33]. Next, the image patch is resized to $224 \times 224$. Finally, we perform normalization on the image by setting both the mean and the standard deviation as $(0.5, 0.5, 0.5)$. As for training BERT and KnowBert, no data augmentation is used other than shuffling the order of scene text before grouping them into a sentence, as both BERT and KnowBert can overfit quickly when the input text is not so abundant.
We adapt AdamW~\cite{iclr/LoshchilovH19} to optimize the whole network with an initial learning rate of 3e-5. The learning rate warmup for 500 iterations and the cosine annealed warm restart strategy are adopted at the same time. All models are trained on the dataset for 10 epochs. 
\begin{figure*}[t]
    \includegraphics[width=0.98\linewidth]{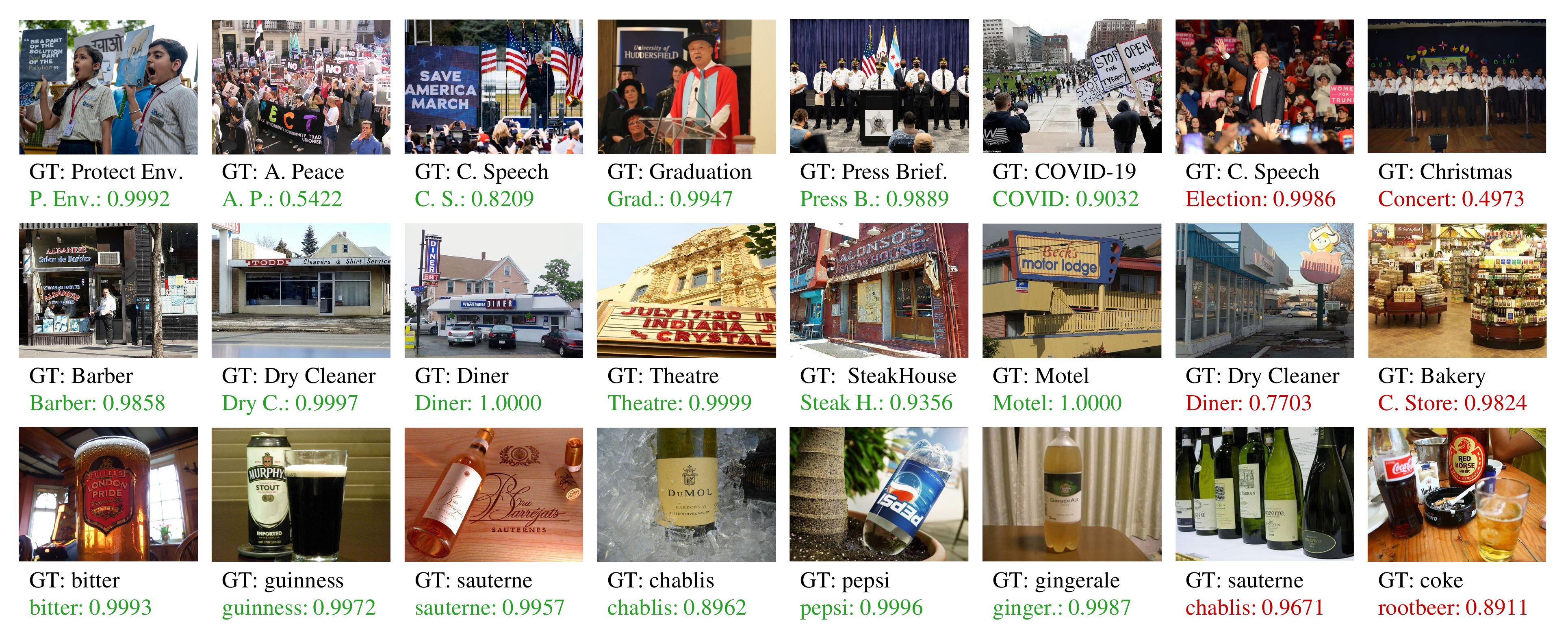}
\centering
% \vspace{-1.5ex}
  \caption{Some examples of classification results. GT denotes Ground Truth. The Top-1 prediction and its probability are shown below each picture. The names of some categories are abbreviated.}
  \label{cls_result}
%   \vspace{-1.5ex}
\end{figure*}

We conduct all experiments based on PyTorch~\cite{NEURIPS2019_9015}. The codes of ResNet-152~\cite{cvpr/HeZRS16} and ViT~\cite{iclr/DosovitskiyB0WZ21} are from \cite{wacv/MaflaDBGK20} and the timm package \cite{rw2019timm}. For both ResNet-152 and ViT, the pre-trained models on ImageNet are used for finetuning.
The implementation of BERT~\cite{naacl/DevlinCLT19} and KnowBert are from the huggingface transformers~\cite{Wolf_Transformers_StateoftheArt_Natural_2020} and~\cite{Peters2019KnowledgeEC}. 
The BookCorpus~\cite{iccv/ZhuKZSUTF15} and English Wikipedia pre-trained model are loaded on BERT. 
In addition, we use~\texttt{torchtext}, which is a package from PyTorch for the GloVe~\cite{PenningtonSM14glove} and fastText~\cite{bojanowski2017enriching}. 

During testing, the shorter side of the image is resized to 224. Then a $224 \times 224$ image patch is cropped from the image center. As for the spotted scene text, we keep their original order for BERT and KnowBert.

\subsection{Baselines on the crowd activity dataset}\label{subsection: baseline on crowd activity}
%We compare our method with several baseline methods including visual baseline (RestNet-152 and ViT), textual/knowledge baseline (fastText and KnowBert), and multi-modal baseline (~\cite{wacv/MaflaDBGK20} and~\cite{wacv/MaflaDBGK21}) on our proposed crowd activity dataset. We conduct two types of experiments using two different dataset setting: 1) The models of visual baseline and multi-modal baseline are trained on all training images and tested on all testing images; 2) The models of textual/knowledge baseline are trained and tested on the subset of images having spotted texts.
We compare our method with several baseline methods, including visual baseline (ResNet-152 and ViT), textual/knowledge baseline (fastText and KnowBert), and multi-modal baseline (~\cite{wacv/MaflaDBGK20}and~\cite{wacv/MaflaDBGK21}) on our proposed crowd activity dataset. We conduct two types of experiments using two different dataset settings. 1) The visual baseline and multi-modal baseline models are trained on all training images and tested on all testing images. 2) The textual/knowledge baseline models are trained and tested on the subset of images consisting of spotted texts. The textual cues used in~\cite{wacv/MaflaDBGK20}and~\cite{wacv/MaflaDBGK21} are from fastText. 

Tab.~\ref{table:baselines on crowd activity} displays the quantitative comparisons on the crowd activity dataset. Among previous methods, ViT achieves state-of-the-art performances, while our method outperforms ViT by 7.2\% mAP. In particular, the improvements of the subset of demonstrations reach more than 11.0\% mAP, which is the highest gain than activities of daily living. The reason is that the visual cues on those demonstration activities are incredibly subtle. For example, most scenarios are that protest marchers hold flags and slogans and walk on the street. Such subtle visual cues require valuable knowledge for better understanding those scenes. Thus, the performance improvement confirms the significance of scene text instances in datasets such as Crowd Activity for the robust classification of fine-grained images.       
%displays the quantitative comparisons on the crowd activity dataset. Among previous methods, Vit achieves the state-of-the-art performances, while our method outperforms ViT by 6.3\% mAP. In particular, the improvements of the subset of demonstrations reach over 10.0\% mAP, which are far from the improvements on those activities of daily living. The reason is that the visual cues on those demonstration activities are extremely subtle. For example, most scenarios are that protest marchers hold flags and slogans and walk on the street. Such subtle visual cues require useful knowledge for better understanding those scenes.
\begin{figure}[t]
    \includegraphics[width=0.85\linewidth]{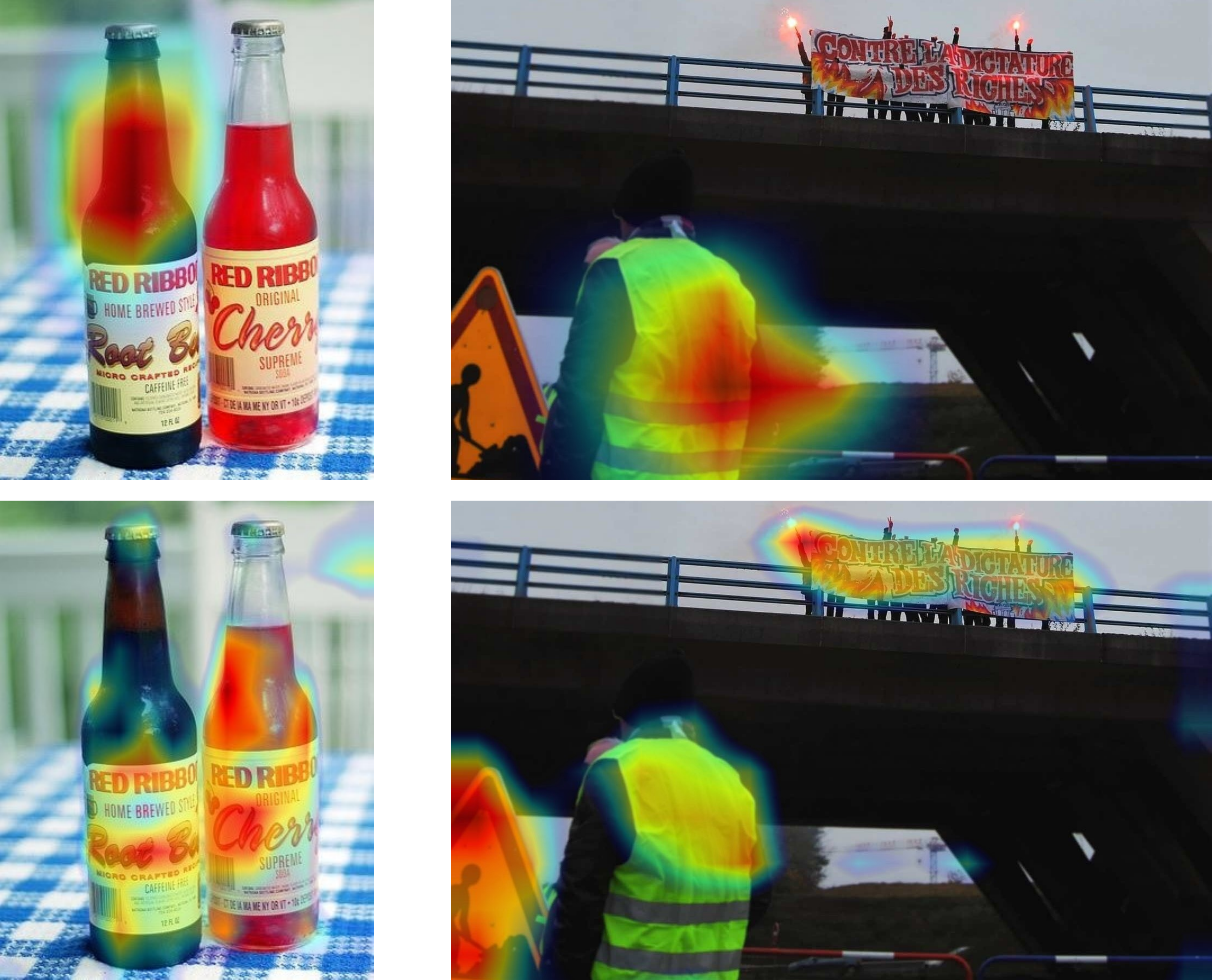}
\centering
%\vspace{-1.5ex}
  \caption{Visualization results. The top two are ResNet-152 grad-CAM~\cite{iccv/SelvarajuCDVPB17} results, and the bottom two are ViT attention maps.}
  \label{heatmap}
 \vspace{-2.5ex}
\end{figure}

\subsection{Comparisons with state-of-the-art method}
Bai~\etal~\cite{access/BaiYLXL18} take GoogLeNet~\cite{cvpr/SzegedyLJSRAEVR15} as visual backbone while the most recent state-of-the-art methods~\cite{wacv/MaflaDBGK20, wacv/MaflaDBGK21} employ ResNet-152~\cite{cvpr/HeZRS16}. For a fair comparison, we first evaluate our method with ResNet-152 and take E2E-MLT~\cite{accv/BustaPM18} as text spotter. Then, we conduct experiments under the setting of ViT and Google OCR.

As shown in Tab.~\ref{sota_result}, our model achieves the best performance on the three datasets. The method~\cite{wacv/MaflaDBGK21} outperforms previous methods by using the features of general objects within images. However, our model surpasses it, by 5.39\% and 3.72\% on Drink Bottle and Con-Text datasets, respectively. The method~\cite{wacv/MaflaDBGK20} does not use the information of general objects. Consequently, our method achieves superior performance on the two public datasets over the method~\cite{wacv/MaflaDBGK20}. The consistent outperformance of our proposed model over existing methods demonstrates the significance and effectiveness of integrating the knowledge behind scene text for better understanding the objects or scene.  
%As shown in Tab. \ref{sota_result}, we can observe that our proposed model achieves the best performance on the three datasets. The method~\cite{wacv/MaflaDBGK21} outperforms previous methods but requires the features of general objects within images. Compared to~\cite{wacv/MaflaDBGK21}, our method still obtains 4.65\% and 2.20\% improvements on Drink Bottle and Con-Text datasets, respectively. For method~\cite{wacv/MaflaDBGK20} that does not use the information of general objects, our method obtains significantly superior performance among the two public datasets. The consistent improvements over existing methods demonstrate that it is essential to introduce knowledge behind scene text for better understanding the objects or scene.  
To further validate the significance of introducing knowledge to this task, we compare our method with~\cite{wacv/MaflaDBGK20} and~\cite{wacv/MaflaDBGK21} on our Crowd Activity dataset. Specifically, we train the model with their officially released codes\footnote{http://github.com/DreadPiratePsyopus/Fine\_Grained\_Clf}\footnote{https://github.com/AndresPMD/GCN\_classification}.
As depicted in Tab.~\ref{sota_result}, our method outperforms the method~\cite{wacv/MaflaDBGK21} by 8.20\% mAP, which further illustrates that mining knowledge is vital to understand the meanings of natural images fully.

As some qualitative results of our method are shown in Fig.~\ref{cls_result}, the proposed method can identify these visually alike images on Drink Bottle and Con-Text datasets. As illustrated in Sec.~\ref{subsection: baseline on crowd activity}, the visual cues and the literal meaning of scene text in images are highly subtle on the crowd activity dataset. Yet, our method still classifies them very well.
\begin{table}[t]
	\centering
	\begin{tabular} {l|ll|ccc}	
	\Xhline{1.0pt}
	& Vision& Emb.& C.T.& Bottles& Activity\\
	\hline
\multirow{2}*{\tabincell{c}{vis.}}
& R152 & - & 70.96 & 73.41 & 71.58 \\
& ViT & - & 79.24 & 80.81 & 80.29 \\
\hline
\multirow{4}*{\tabincell{c}{vis. + \\ text}}
& R152 & GloVe & 73.97 & 76.67 & 74.75 \\
& R152 & fastText & 73.66 & 76.67 & 74.89 \\
& ViT & GloVe & 79.79 & 80.56 & 81.25 \\
& ViT & fastText & 79.82 & 81.18 & 80.71 \\
\hline
\multirow{4}*{\tabincell{c}{vis. + \\ text + \\ know.}}
% & R152 & BERT & 83.15 & 79.58 & 82.68 \\
& R152 & BERT & 81.59 & 77.94 & 81.68 \\
% & R152 & KB & 84.51 & 79.32 & 83.79 \\
& R152 & KB & 85.42 & 80.17 & 83.79 \\
& ViT & BERT & 86.51 & 82.81 & 85.34 \\
% & ViT & KB & \textbf{88.34} & \textbf{85.26} & \textbf{87.45} \\
& ViT & KB & \textbf{89.53} & \textbf{85.26} & \textbf{87.45} \\
	\Xhline{1.0pt}
	\end{tabular}
	\vspace{-0.8ex}
	\caption{Performances of different vision and embedding models combinations on three datasets. R152 denotes ResNet-152. The abbreviated names, vis. and know., mean visual and knowledge. The text + know. means the features containing knowledge of texts. C.T. means Con-Text. (Metric: mAP)}
	\label{ablation_map}
	\vspace{-1.5ex}
\end{table}

\subsection{Ablation study}
This section provides detailed ablation studies to validate the effect of different modules included in the proposed model for mining knowledge. Thus, we present the performances on the three datasets under various combinations of visual features and textual features. Then, we discuss the impact of KARC and VKAC components. Finally, we show the advantage of jointly optimizing the whole network. 
% Note that the default scene text reading system for this section is Google OCR engineering.

%In this section, we first provide elaborate ablation studies to further verify the effectiveness of mining knowledge to this task. Thus, we give the performances on the three datasets under different combinations of visual features (~\ie, ResNet-152 and ViT) and textual features (~\ie, Glove, fastText, BERT, and KnowBert). 
%Then, we show the advantage of jointly optimizing the whole network. The scene text reading system that we employ in this section is Google OCR engineering.

\textbf{The impact of visual features} As shown in Tab.~\ref{ablation_map}, introducing textual cues (~\ie, Glove and fastText) to the ResNet-152 model can significantly improve the performance up to 3\% mAP. However, ViT model performance improvement is not more than 1\% mAP. We further compare the two models with qualitative examples via visualizing the attention map of both models (only trained with image data) of ResNet-152 and ViT. As depicted in Fig.~\ref{heatmap}, the ResNet-152 model mainly focuses on the visual contents.  However, the ViT model captures the visual contents and harvest the textual cues from the image by self-attention mechanism. Thus, embedding features provides complementary information to boost the performances of ViT instead of solely exploiting the literal meaning of scene text.
%As shown in Tab.~\ref{ablation_map}, introducing textual cues (~\ie, Glove and fastText) to the ResNet-152 model can significantly improve the performance by around 3\% mAP. However, this strategy only achieves at most 1\% mAP improvements to the ViT model. To explain this phenomenon, we visualize the attention map of both models (only trained with image data) of ResNet-152 and ViT. As depicted in Fig.~\ref{heatmap}, we find that the model of ResNet-152 mainly focuses on the visual contents, while ViT can capture the visual contents and harvest the textual cues from the image by its self-attention mechanism. That is why solely exploiting the literal meaning of scene text hardly boosts the performances of ViT, as these embedding features provide little complementary information to ViT.

\begin{table}[t]
	\centering
	\begin{tabular}{l|cc|ccc}
	\Xhline{1.0pt}
	&KARC& VKAC& C.T.& Bottles& Activity\\
	\hline
    Baseline&        &         & 86.51 & 82.81 & 85.34 \\
    model A& $\surd$ &         & 87.25 & 83.59 & 86.16 \\
    model B& $\surd$ & $\surd$ & \textbf{89.53} & \textbf{85.26} & \textbf{87.45} \\
	\Xhline{1.0pt}
	\end{tabular}
	\vspace{-0.8ex}
	\caption{Ablation studies of KARC and VKAC components. ViT is applied to extract features from images. C.T. denotes the Con-Text dataset. (Metric: mAP)}
	\label{ablation_components}
	\vspace{-1.5ex}
\end{table}

% \begin{table}[t]
% 	\centering
% 	\begin{tabular}{cc|ccc}
% 	\Xhline{1.0pt}
% 	KARC& VKAC& Con-Text& Bottles& Activity\\
% 	\hline
%             &         & 86.51 & 82.81 & 85.34 \\
%      $\surd$ &         & 87.25 & 83.59 & 86.16 \\
%      $\surd$ & $\surd$ & \textbf{89.53} & \textbf{85.26} & \textbf{87.45} \\
% 	\Xhline{1.0pt}
% 	\end{tabular}
% 	\vspace{-0.8ex}
% 	\caption{Ablation studies of KARC and VKAC components. ViT is applied to extract features from images. (Metric: mAP)}
% 	\label{ablation_components}
% 	\vspace{-1.5ex}
% \end{table}

\begin{figure}[t]
    \includegraphics[width=0.95\linewidth]{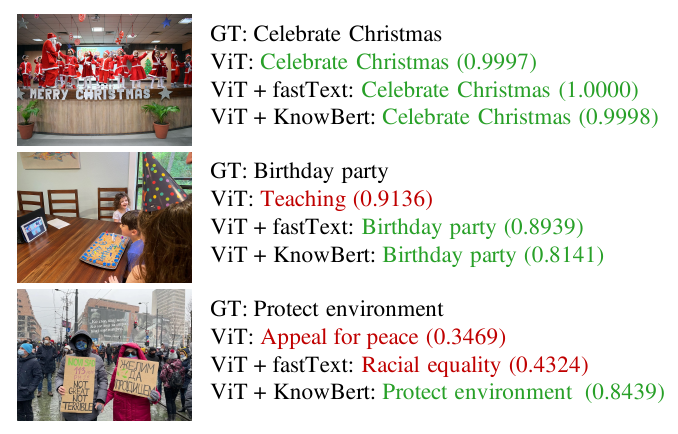}
\centering
\vspace{-1.5ex}
  \caption{The classification results of different models.}
  \label{cls_result2}
 \vspace{-2.5ex}
\end{figure}

\begin{table}[t]
	\centering
	\begin{tabular}{l|ccc}
	\Xhline{1.0pt}
	Model & Con-Text & Bottles & Activity \\
	\hline
    ViT & 79.24 & 80.81 & 80.29 \\
    KnowBert & 47.07 & 53.28 & 64.66 \\
    model A & 81.47 & 82.26 & 81.79 \\
    model B & \textbf{89.53} & \textbf{85.26} & \textbf{87.45} \\
	\Xhline{1.0pt}
	\end{tabular}
	\vspace{-0.8ex}
	\caption{The model A is trained in a separated manner. The model B is trained in an end-to-end manner. (Metric: mAP)}
	\label{comparison}
	\vspace{-2.5ex}
\end{table}
\textbf{The impact of knowledge-enhanced features} 
As mentioned before, a direct way to mine knowledge is to exploit the BERT encoder output features. As shown in Tab.~\ref{ablation_map}, the employment of knowledge-enhanced features from BERT achieves significant improvements than the typical word embedding features (GloVe/fastText). The ViT+BERT model surpasses the performance of the ViT+fastText model by 6.69\%, 1.63\%, 4.63\% on Con-Text, Drink Bottle, and Crowd Activity. This superior performance proves that the explicit knowledge in knowledge bases significantly enriches the semantics of scene text for understanding objects. Furthermore, unlike BERT, KnowBert explicitly introduces knowledge from a knowledge base into the model. The experimental results show that the KnowBert model consistently outperforms the BERT model. Therefore, introducing knowledge behind scene text to neural network feature learning enhances understanding natural images. As shown in Fig.~\ref{cls_result2}, the employment of knowledge substantially enriches the classification accuracy, as the knowledge behind ``PM2.5" tells that the third image is about environment.
%As mentioned before, a direct way to mine knowledge is to exploit the outputted features of the encoder of BERT. As shown in Tab.~\ref{ablation_map}, we can observe that the usage of knowledge-enhanced features from BERT achieves significant improvements than the typical word embedding features from GloVe and fastText. The ViT+BERT model improves the performance by 6.69\%, 1.63\%, 4.63\% comparing to the ViT+fastText model on Con-Text, Drink Bottle, and Crowd Activity datasets, respectively. Therefore, we can conclude that introducing knowledge behind scene text can better assist the understanding of natural images than solely relying on its literal meaning. Furthermore, unlike BERT that implicitly contains knowledge, KnowBert explicitly introduces knowledge from a knowledge base into the BERT model. From our experiment, KnowBert consistently obtains better performance than the BERT model. Those experimental results demonstrate that the explicit knowledge in knowledge bases can further enrich the semantics of scene text for understanding objects. As shown in Fig.~\ref{cls_result2}, the usage of knowledge could understand the scene better. 

\textbf{The impact of KARC and VKAC components} As shown in Tab.~\ref{ablation_components} Model B is the default model equipped with KARC and VKAC, while model A only employed KARC. The integration of KARC only in model A improves the performance on all datasets. Moreover, integrating VKAC on top of KARC in model B increases the recognition performance mAP by 2.28\%, 1.67\%, and 1.29\% on Con-Text, Bottles, and Crowd Activity datasets, respectively. The experimental results demonstrate the effectiveness of fusing multi-modal features for this task.

%The performances are improved on all datasets by introducing KARC.
%Moreover, VKAC that attends to very pertinent knowledge for the objects further increases the mAP by 2.28\%, 1.67\%, and 1.29\% on Con-Text, Bottles, and Crowd Activity datasets. 
%The experimental results demonstrate the effectiveness of fusing multi-modal features for this task. 

\textbf{Joint optimization} Integrating the process of mining knowledge, feature extraction, and classification in a unified network makes it feasible to optimize them jointly. The model that is jointly optimized could achieve better performance than the one with separated feature extraction and classifier, as those processes are complementary to each other. To confirm this assumption, we first train the models of ViT and KnowBert with image data and scene text, respectively, at the supervision of the classification task. Then, the classifier and VKAC are trained, accepting as input visual features and knowledge-enhanced features extracted from the pre-trained models. As reported in Tab.~\ref{comparison}, the model trained in an end-to-end manner significantly outperforms the one trained separately, showing the necessity of integrating knowledge mining process into the network.  

% Furthermore, we report the performance under the metric of accuracy in Tab.~\ref{ablation_acc}. The experimental results consistently validate the effectiveness of mining knowledge for this task. In addition, the performances of solely using the visual features is far from the ones on Con-Text and Bottles datasets, which indicates the demands on more precise semantics of objects for crowd activity recognition.

% \begin{table}[t]
% 	\centering
% 	\begin{tabular}{llccc}
% 	\Xhline{1.0pt}
% 	Vision& Emb.& Con-Text& Bottles& Activity\\
% 	\hline
% R152 & None & 66.47 & 68.75 & 59.90 \\
% R152 & GloVe & 70.28 & 72.55 & 63.08 \\
% R152 & fastText & 69.64 & 72.46 & 63.25 \\
% R152 & BERT & 78.89 & 76.31 & 73.47 \\
% R152 & KnowBert & 79.72 & 75.30 & 76.29 \\
% ViT & None & 76.39 & 76.33 & 70.85 \\
% ViT & GloVe & 76.49 & 76.76 & 72.01 \\
% ViT & fastText & 76.60 & 77.25 & 71.48 \\
% ViT & BERT & 84.67 & 79.75 & 79.77 \\
% ViT & KnowBert & \textbf{85.15} & \textbf{81.24} & \textbf{81.19} \\
% 	\Xhline{1.0pt}
% 	\end{tabular}
% % 	\vspace{+0.8ex}
% 	\caption{Performances of different vision and embedding models combinations on three datasets. R152 denotes ResNet-152. (Metric: Accuracy/\%)}
% 	\label{ablation_acc}
% % 	\vspace{-2ex}
% \end{table}

% \FloatBarrier
\section{Conclusion}\label{conclusion}

In this paper, we have confirmed that the usage of the knowledge behind scene text can improve the performance of the fine-grained image classification task. Experiments on the two benchmark datasets and the proposed Crowd Activity dataset have verified the effectiveness and efficiency of our method for product recognition and crowd activity analysis. In the future, we will further explore the usage of knowledge mining of scene text on other tasks of multi-modal fusion, such as scene text, visual question and answering, and visual grounding.

% \section{Acknowledgements}
\noindent\textbf{Acknowledgements} This work was supported by the National Natural Science Foundation of China 61733007.

%%%%%%%%% REFERENCES
{\small
\bibliographystyle{ieee_fullname}
\bibliography{egbib}

\begin{thebibliography}{10}\itemsep=-1pt

\bibitem{pami/AlmazanGFV14}
Jon Almaz{\'{a}}n, Albert Gordo, Alicia Forn{\'{e}}s, and Ernest Valveny.
\newblock Word spotting and recognition with embedded attributes.
\newblock {\em {IEEE} TPAMI}, 2014.

\bibitem{access/BaiYLXL18}
Xiang Bai, Mingkun Yang, Pengyuan Lyu, Yongchao Xu, and Jiebo Luo.
\newblock Integrating scene text and visual appearance for fine-grained image
  classification.
\newblock {\em {IEEE} Access}, 6:66322--66335, 2018.

\bibitem{iccv/BitenTMBRJVK19}
Ali~Furkan Biten, Rub{\`{e}}n Tito, Andr{\'{e}}s Mafla,
  Llu{\'{\i}}s~G{\'{o}}mez i Bigorda, Mar{\c{c}}al Rusi{\~{n}}ol, C.~V.
  Jawahar, Ernest Valveny, and Dimosthenis Karatzas.
\newblock Scene text visual question answering.
\newblock In {\em ICCV}, pages 4290--4300, 2019.

\bibitem{bojanowski2017enriching}
Piotr Bojanowski, Edouard Grave, Armand Joulin, and Tomas Mikolov.
\newblock Enriching word vectors with subword information.
\newblock {\em {TACL}}, 2017.

\bibitem{accv/BustaPM18}
Michal Busta, Yash Patel, and Jiri Matas.
\newblock {E2E-MLT} - an unconstrained end-to-end method for multi-language
  scene text.
\newblock In Gustavo Carneiro and Shaodi You, editors, {\em ACCV}, 2018.

\bibitem{Cordonnier2021CVPR}
Jean-Baptiste Cordonnier, Aravindh Mahendran, Alexey Dosovitskiy, Dirk
  Weissenborn, Jakob Uszkoreit, and Thomas Unterthiner.
\newblock Differentiable patch selection for image recognition.
\newblock In {\em CVPR}, pages 2351--2360, 2021.

\bibitem{cvpr/DengDSLL009}
Jia Deng, Wei Dong, Richard Socher, Li{-}Jia Li, Kai Li, and Fei{-}Fei Li.
\newblock Imagenet: {A} large-scale hierarchical image database.
\newblock In {\em {CVPR}}, 2009.

\bibitem{naacl/DevlinCLT19}
Jacob Devlin, Ming{-}Wei Chang, Kenton Lee, and Kristina Toutanova.
\newblock {BERT:} pre-training of deep bidirectional transformers for language
  understanding.
\newblock In {\em {NAACL-HLT}}, 2019.

\bibitem{iclr/DosovitskiyB0WZ21}
Alexey Dosovitskiy, Lucas Beyer, Alexander Kolesnikov, Dirk Weissenborn,
  Xiaohua Zhai, Thomas Unterthiner, Mostafa Dehghani, Matthias Minderer, Georg
  Heigold, Sylvain Gelly, Jakob Uszkoreit, and Neil Houlsby.
\newblock An image is worth 16x16 words: Transformers for image recognition at
  scale.
\newblock In {\em {ICLR}}, 2021.

\bibitem{cvpr/FuZM17}
Jianlong Fu, Heliang Zheng, and Tao Mei.
\newblock Look closer to see better: Recurrent attention convolutional neural
  network for fine-grained image recognition.
\newblock In {\em CVPR}, 2017.

\bibitem{emnlp/GaneaH17}
Octavian{-}Eugen Ganea and Thomas Hofmann.
\newblock Deep joint entity disambiguation with local neural attention.
\newblock In {\em EMNLP}, 2017.

\bibitem{dicta/GeMSWLRC16}
ZongYuan Ge, Chris McCool, Conrad Sanderson, Peng Wang, Lingqiao Liu, Ian~D.
  Reid, and Peter~I. Corke.
\newblock Exploiting temporal information for dcnn-based fine-grained object
  classification.
\newblock In {\em {DICTA}}, 2016.

\bibitem{cvpr/HeZRS16}
Kaiming He, Xiangyu Zhang, Shaoqing Ren, and Jian Sun.
\newblock Deep residual learning for image recognition.
\newblock In {\em CVPR}, 2016.

\bibitem{emnlp/HoffartYBFPSTTW11}
Johannes Hoffart, Mohamed~Amir Yosef, Ilaria Bordino, Hagen F{\"{u}}rstenau,
  Manfred Pinkal, Marc Spaniol, Bilyana Taneva, Stefan Thater, and Gerhard
  Weikum.
\newblock Robust disambiguation of named entities in text.
\newblock In {\em EMNLP}, 2011.

\bibitem{tmm/KaraogluTGS17}
Sezer Karaoglu, Ran Tao, Theo Gevers, and Arnold W.~M. Smeulders.
\newblock Words matter: Scene text for image classification and retrieval.
\newblock {\em {IEEE} TMM}, 2017.

\bibitem{mm/KaraogluGG13}
Sezer Karaoglu, Jan~C. van Gemert, ., and Theo Gevers.
\newblock Con-text: text detection using background connectivity for
  fine-grained object classification.
\newblock In {\em {ACM} MM}, 2013.

\bibitem{emnlp/KhanMLS20}
Aisha~Urooj Khan, Amir Mazaheri, Niels da Vitoria~Lobo, and Mubarak Shah.
\newblock {MMFT-BERT:} multimodal fusion transformer with {BERT} encodings for
  visual question answering.
\newblock In {\em {EMNLP}}, 2020.

\bibitem{khosla2011novel}
Aditya Khosla, Nityananda Jayadevaprakash, Bangpeng Yao, and Fei-Fei Li.
\newblock Novel dataset for fine-grained image categorization: Stanford dogs.
\newblock In {\em Proc. CVPR Workshop on FGVC}, 2011.

\bibitem{iccvw/Krause0DF13}
Jonathan Krause, Michael Stark, Jia Deng, and Li Fei{-}Fei.
\newblock 3d object representations for fine-grained categorization.
\newblock In {\em ICCVW}, 2013.

\bibitem{eccv/LiaoPHHB20}
Minghui Liao, Guan Pang, Jing Huang, Tal Hassner, and Xiang Bai.
\newblock Mask textspotter v3: Segmentation proposal network for robust scene
  text spotting.
\newblock In {\em {ECCV}}, 2020.

\bibitem{iclr/LoshchilovH19}
Ilya Loshchilov and Frank Hutter.
\newblock Decoupled weight decay regularization.
\newblock In {\em {ICLR}}, 2019.

\bibitem{wacv/MaflaDBGK20}
Andr{\'{e}}s Mafla, Sounak Dey, Ali~Furkan Biten, Llu{\'{\i}}s G{\'{o}}mez, and
  Dimosthenis Karatzas.
\newblock Fine-grained image classification and retrieval by combining visual
  and locally pooled textual features.
\newblock In {\em {WACV}}, 2020.

\bibitem{wacv/MaflaDBGK21}
Andr{\'{e}}s Mafla, Sounak Dey, Ali~Furkan Biten, Llu{\'{\i}}s G{\'{o}}mez, and
  Dimosthenis Karatzas.
\newblock Multi-modal reasoning graph for scene-text based fine-grained image
  classification and retrieval.
\newblock In {\em {WACV}}, 2021.

\bibitem{corr/MajiRKBV13}
Subhransu Maji, Esa Rahtu, Juho Kannala, Matthew~B. Blaschko, and Andrea
  Vedaldi.
\newblock Fine-grained visual classification of aircraft.
\newblock {\em CoRR}, 2013.

\bibitem{cacm/Miller95}
George~A. Miller.
\newblock Wordnet: {A} lexical database for english.
\newblock {\em Commun. {ACM}}, 1995.

\bibitem{cvpr/Movshovitz-Attias15}
Yair Movshovitz{-}Attias, Qian Yu, Martin~C. Stumpe, Vinay~D. Shet, Sacha
  Arnoud, and Liron Yatziv.
\newblock Ontological supervision for fine grained classification of street
  view storefronts.
\newblock In {\em CVPR}, 2015.

\bibitem{NEURIPS2019_9015}
Adam Paszke, Sam Gross, Francisco Massa, Adam Lerer, James Bradbury, Gregory
  Chanan, Trevor Killeen, Zeming Lin, Natalia Gimelshein, Luca Antiga, Alban
  Desmaison, Andreas Kopf, Edward Yang, Zachary DeVito, Martin Raison, Alykhan
  Tejani, Sasank Chilamkurthy, Benoit Steiner, Lu Fang, Junjie Bai, and Soumith
  Chintala.
\newblock Pytorch: An imperative style, high-performance deep learning library.
\newblock In {\em NeurIPS}. 2019.

\bibitem{PenningtonSM14glove}
Jeffrey Pennington, Richard Socher, and Christopher~D. Manning.
\newblock Glove: Global vectors for word representation.
\newblock In {\em {EMNLP}}, 2014.

\bibitem{cvpr/PerronninD07}
Florent Perronnin and Christopher~R. Dance.
\newblock Fisher kernels on visual vocabularies for image categorization.
\newblock In {\em {CVPR}}, 2007.

\bibitem{naacl/PetersNIGCLZ18}
Matthew~E. Peters, Mark Neumann, Mohit Iyyer, Matt Gardner, Christopher Clark,
  Kenton Lee, and Luke Zettlemoyer.
\newblock Deep contextualized word representations.
\newblock In {\em {NAACL-HLT}}, 2018.

\bibitem{Peters2019KnowledgeEC}
Matthew~E. Peters, Mark Neumann, Robert~L Logan, Roy Schwartz, Vidur Joshi,
  Sameer Singh, and Noah~A. Smith.
\newblock Knowledge enhanced contextual word representations.
\newblock In {\em EMNLP}, 2019.

\bibitem{petroni2019language}
Fabio Petroni, Tim Rockt{\"a}schel, Sebastian Riedel, Patrick Lewis, Anton
  Bakhtin, Yuxiang Wu, and Alexander Miller.
\newblock Language models as knowledge bases?
\newblock In {\em EMNLP-IJCNLP}, 2019.

\bibitem{iccv/PlummerWCCHL15}
Bryan~A. Plummer, Liwei Wang, Chris~M. Cervantes, Juan~C. Caicedo, Julia
  Hockenmaier, and Svetlana Lazebnik.
\newblock Flickr30k entities: Collecting region-to-phrase correspondences for
  richer image-to-sentence models.
\newblock In {\em ICCV}, 2015.

\bibitem{corr/abs-2007-00655}
Corby Rosset, Chenyan Xiong, Minh Phan, Xia Song, Paul~N. Bennett, and Saurabh
  Tiwary.
\newblock Knowledge-aware language model pretraining.
\newblock {\em CoRR}, 2020.

\bibitem{iccv/SelvarajuCDVPB17}
Ramprasaath~R. Selvaraju, Michael Cogswell, Abhishek Das, Ramakrishna Vedantam,
  Devi Parikh, and Dhruv Batra.
\newblock Grad-cam: Visual explanations from deep networks via gradient-based
  localization.
\newblock In {\em {ICCV}}, 2017.

\bibitem{textvqaknow}
Ajeet~Kumar Singh, Anand Mishra, Shashank Shekhar, and Anirban Chakraborty.
\newblock From strings to things: Knowledge-enabled vqa model that can read and
  reason.
\newblock In {\em ICCV}, 2019.

\bibitem{lrec/SpitkovskyC12}
Valentin~I. Spitkovsky and Angel~X. Chang.
\newblock A cross-lingual dictionary for english wikipedia concepts.
\newblock In {\em LREC}, 2012.

\bibitem{cvpr/SzegedyLJSRAEVR15}
Christian Szegedy, Wei Liu, Yangqing Jia, Pierre Sermanet, Scott~E. Reed,
  Dragomir Anguelov, Dumitru Erhan, Vincent Vanhoucke, and Andrew Rabinovich.
\newblock Going deeper with convolutions.
\newblock In {\em {CVPR}}, 2015.

\bibitem{Wang_2021_CVPR}
Hao Wang, Xiang Bai, Mingkun Yang, Shenggao Zhu, Jing Wang, and Wenyu Liu.
\newblock Scene text retrieval via joint text detection and similarity
  learning.
\newblock In {\em CVPR}, 2021.

\bibitem{aaai/WangLZYBXHW020}
Hao Wang, Pu Lu, Hui Zhang, Mingkun Yang, Xiang Bai, Yongchao Xu, Mengchao He,
  Yongpan Wang, and Wenyu Liu.
\newblock All you need is boundary: Toward arbitrary-shaped text spotting.
\newblock In {\em {AAAI}}, 2020.

\bibitem{rw2019timm}
Ross Wightman.
\newblock Pytorch image models.
\newblock \url{https://github.com/rwightman/pytorch-image-models}, 2019.

\bibitem{Wolf_Transformers_StateoftheArt_Natural_2020}
Thomas Wolf, Lysandre Debut, Victor Sanh, Julien Chaumond, Clement Delangue,
  Anthony Moi, Pierric Cistac, Tim Rault, R{\'{e}}mi Louf, Morgan Funtowicz,
  Joe Davison, Sam Shleifer, Patrick von Platen, Clara Ma, Yacine Jernite,
  Julien Plu, Canwen Xu, Teven~Le Scao, Sylvain Gugger, Mariama Drame, Quentin
  Lhoest, and Alexander~M. Rush.
\newblock Transformers: State-of-the-art natural language processing.
\newblock In Qun Liu and David Schlangen, editors, {\em EMNLP}, 2020.

\bibitem{iccv/ZhuKZSUTF15}
Yukun Zhu, Ryan Kiros, Richard~S. Zemel, Ruslan Salakhutdinov, Raquel Urtasun,
  Antonio Torralba, and Sanja Fidler.
\newblock Aligning books and movies: Towards story-like visual explanations by
  watching movies and reading books.
\newblock In {\em {ICCV}}, 2015.

\end{thebibliography}
}

\end{document}